\documentclass[conference]{IEEEtran}
\IEEEoverridecommandlockouts
\usepackage{cite}
\usepackage{amsmath,amssymb,amsfonts}
\usepackage{algorithmic}
\usepackage{graphicx}
\usepackage{textcomp}
\usepackage{xcolor}
\usepackage{float}
\usepackage{tabularx}
\usepackage{multirow}
\usepackage{balance}

\def\BibTeX{{\rm B\kern-.05em{\sc i\kern-.025em b}\kern-.08em
    T\kern-.1667em\lower.7ex\hbox{E}\kern-.125emX}}
\begin{document}

\title{Deepfake Forensic Analysis: Source Dataset Attribution and Legal Implications of Synthetic Media Manipulation}

\author{
    \IEEEauthorblockN{
        Massimiliano Cassia\IEEEauthorrefmark{1}, 
        Luca Guarnera\IEEEauthorrefmark{2}, 
        Mirko Casu\IEEEauthorrefmark{3}, 
        Ignazio Zangara\IEEEauthorrefmark{4}, 
        Sebastiano Battiato\IEEEauthorrefmark{5}
    }
    \IEEEauthorblockA{
        \textit{Department of Mathematics and Computer Science} \\
        \textit{University of Catania}\\
        Catania, Italy \\
        \IEEEauthorrefmark{1}\texttt{cssmsm01b07i754z@studium.unict.it} \\
        \IEEEauthorrefmark{2}\texttt{luca.guarnera@unict.it}, ORCID: \texttt{0000-0001-8315-351X} \\
        \IEEEauthorrefmark{3}\texttt{mirko.casu@phd.unict.it}, ORCID: \texttt{0000-0001-6975-2241} \\
        \IEEEauthorrefmark{4}\texttt{izangara@lex.unict.it} \\
        \IEEEauthorrefmark{5}\texttt{sebastiano.battiato@unict.it}, ORCID: \texttt{0000-0001-6127-2470}
    }
}

\maketitle

\begin{abstract}
Synthetic media generated by Generative Adversarial Networks (GANs) pose significant challenges in verifying authenticity and tracing dataset origins, raising critical concerns in copyright enforcement, privacy protection, and legal compliance. This paper introduces a novel forensic framework for identifying the training dataset (e.g., CelebA or FFHQ) of GAN-generated images through interpretable feature analysis. By integrating spectral transforms (Fourier/DCT), color distribution metrics, and local feature descriptors (SIFT), our pipeline extracts discriminative statistical signatures embedded in synthetic outputs. Supervised classifiers (Random Forest, SVM, XGBoost) achieve 98–99\% accuracy in binary classification (real vs. synthetic) and multi-class dataset attribution across diverse GAN architectures (StyleGAN, AttGAN, GDWCT, StarGAN, and StyleGAN2). Experimental results highlight the dominance of frequency-domain features (DCT/FFT) in capturing dataset-specific artifacts, such as upsampling patterns and spectral irregularities, while color histograms reveal implicit regularization strategies in GAN training. We further examine legal and ethical implications, showing how dataset attribution can address copyright infringement, unauthorized use of personal data, and regulatory compliance under frameworks like GDPR and California’s AB 602. Our framework advances accountability and governance in generative modeling, with applications in digital forensics, content moderation, and intellectual property litigation.
\end{abstract}

\begin{IEEEkeywords}
deepfake forensics, source dataset attribution, GAN fingerprinting, frequency analysis, DCT analysis, Fourier analysis, digital forensics
\end{IEEEkeywords}

\section{Introduction}

The rapid advancement of digital manipulation technologies has transformed the creation and dissemination of synthetic media, fundamentally altering the landscape of digital content generation \cite{Whittaker2020All, Zia2024Improving}. At the forefront of these innovations, Generative Adversarial Networks (GANs) \cite{goodfellow2014generative} have emerged as a powerful tool for producing highly realistic synthetic images and videos. While these technologies unlock immense creative and technical possibilities, they also pose significant challenges, particularly regarding the origins \cite{brown2020} and ethical use of training datasets \cite{crawford2021}. The remarkable realism of GAN-generated content raises pressing questions about the provenance of datasets—such as CelebA or FFHQ—and the associated risks of copyright infringement, privacy violations, and ethical concerns \cite{amerini2025}. In this context, Casu et al. \cite{Casu2024GenAI} introduced the concept of “impostor bias,” describing a growing distrust in multimedia authenticity driven by the sophistication of AI-generated content, which further complicates forensic decision-making and deepfake detection.

These issues have become increasingly urgent as the digital forensics and legal communities confront the implications of synthetic media, including intellectual property disputes, unauthorized use of personal images, and breaches of data protection regulations \cite{Mirsky2020The, Ghodke2024Ethical}. Traditional forensic techniques—such as analyzing physical evidence, metadata, or visual cues—are often inadequate for addressing the complexity of modern synthetic media and its underlying datasets \cite{Verdoliva2020Media, Corvi2022On}. To bridge this gap, Pontorno et al. \cite{Pontorno2024DeepFeatureX} proposed DeepFeatureX Net, a novel block-based architecture that extracts discriminative features to distinguish diffusion-model-generated, GAN-generated, and real images. This approach achieved state-of-the-art generalization, even under JPEG compression and various adversarial attacks, using intentionally unbalanced datasets. Others \cite{Guarnera2022DeepfakeStyleTransferMixture} have begun exploring forensic ballistics on deepfakes subject to style-transfer pipelines, demonstrating that repeated generative processing leaves distinctive artifacts that can be traced back to specific style-transfer chains.

This work, at the intersection of computer vision, digital forensics, and legal technology, proposes a preliminary framework to infer generative model training datasets by analyzing visual, statistical, and architectural signatures in synthetic images. Unique dataset and GAN configuration artifacts enable tracing dataset provenance, detecting copyright infringements, and addressing ethical concerns. This approach fosters robust, interpretable dataset tracing tools to tackle legal and governance challenges of synthetic media, advancing digital forensics and synthetic media analysis through two core contributions:  
    
\begin{itemize}
    \item \textbf{Technical Innovation:} Develops a novel pipeline for source training dataset attribution combining spectral analysis (Fourier/DCT transforms), color distribution histograms, and local feature descriptors (SIFT) to extract interpretable statistical signatures from deepfake images.
    \item \textbf{Legal and Ethical Exploration:} Investigates dataset attribution as a tool for addressing copyright enforcement, privacy violations, and regulatory compliance, bridging technical analysis with real-world governance challenges posed by synthetic media.
\end{itemize}

The remainder of this paper is organized as follows. Section \ref{sec:sota} reviews related work in deepfake detection and dataset attribution. 
Section \ref{sec:met} details our methodology, including dataset construction, feature extraction, and classification strategies. Section \ref{sec:res} presents experimental results. 
Section \ref{sec:legal} examines the legal and ethical implications of dataset tracing in synthetic media governance. Finally, Section \ref{sec:conclusion} concludes with limitations, future directions, and broader implications for digital forensics and AI ethics.

\section{Related Work}
\label{sec:sota}
The detection and attribution of deepfake content has garnered increasing attention in recent years \cite{khoo2022deepfake}. Traditional detection methods often rely on temporal inconsistencies, such as blinking patterns \cite{li2018ictu} or lip-sync errors \cite{jia2024}, which are effective for video but less so for static images. Other approaches exploit physiological anomalies \cite{hernandez2020} or metadata inconsistencies, but these methods often fail when synthetic content is generated with high fidelity.
Recent studies have shifted focus toward analyzing the intrinsic statistical characteristics embedded in synthetic images \cite{liu2021}. Some techniques utilize deep neural networks to learn subtle artifacts introduced by GANs \cite{afchar2018,yu2019}, while others apply handcrafted features such as frequency transforms \cite{liu2021} or image quality metrics. For instance, Verdoliva \cite{Verdoliva2020Media} emphasized media forensics through noise residual analysis, and Corvi et al. \cite{corvi2024continuous} demonstrated promising results using spectral fingerprints for diffusion-based models.

However, few works specifically target the problem of source dataset attribution—a gap this paper aims to fill. Most existing literature either addresses deepfake detection \cite{yang2022deepfake,guarnera2024mastering} as a binary classification problem or focuses on identifying the generative model architecture. Our approach, by contrast, isolates the influence of the training dataset itself, regardless of the model used. While proactive methods like artificial fingerprinting \cite{yu2021artificial} embed identifiers into training data, our passive method leverages intrinsic features for attribution. Additionally, dataset inference techniques \cite{maini2021dataset} aim to resolve ownership by identifying whether a model was trained on a specific dataset, but our method directly attributes the generated content to its training dataset using interpretable features.

Moreover, while methods such as PRNU (Photo-Response Non-Uniformity) \cite{lukas2006} and image hash comparison \cite{venkatesan2000} have been used for source camera identification, these are not directly applicable to synthetic media \cite{cozzolino2021}. Instead, our use of Discrete Cosine Transform (DCT) and Fast Fourier
Transform (FFT) \cite{frank2020}, color histograms \cite{wang2022}, and Scale-Invariant Feature Transform (SIFT) features \cite{yang2019} in combination represents a novel, interpretable, and domain-agnostic solution for deepfake provenance analysis, applicable across different GAN architectures and real-world forensic scenarios.

\section{Methodology}
\label{sec:met}


\begin{figure}
    \centering
    \includegraphics[width=1\linewidth]{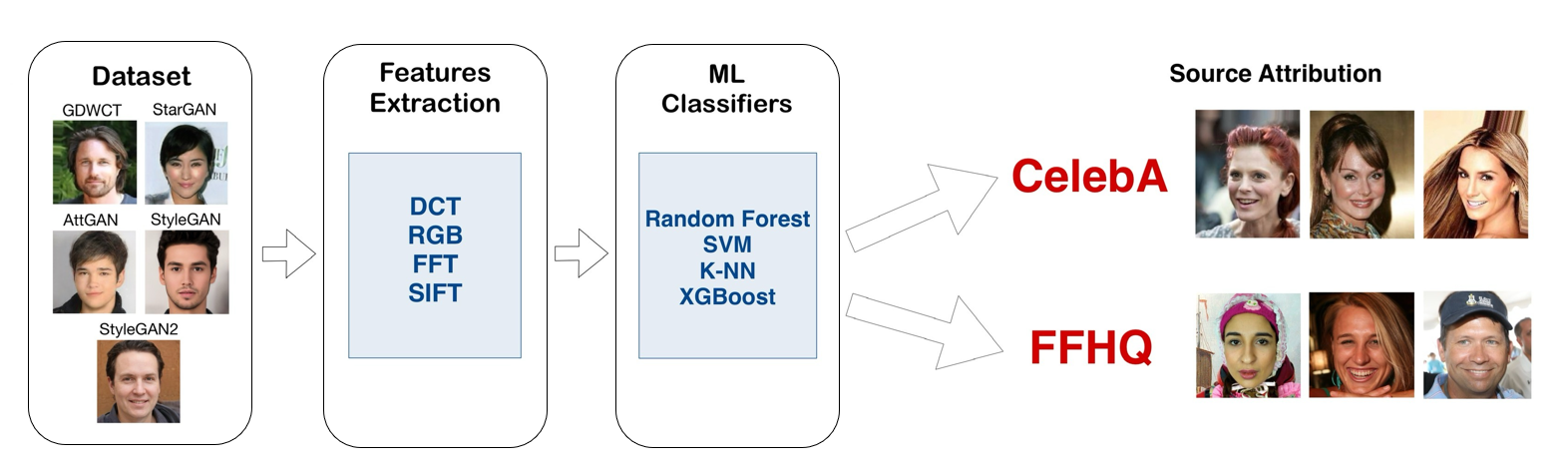}
    \caption{Methodological Pipeline for Dataset Attribution Using Multimodal Features and Machine Learning Classifiers. The pipeline illustrates the process from dataset selection (including GAN-generated images from CelebA and FFHQ) to feature extraction (using DCT, RGB histograms, FFT, and SIFT) and classification using Random Forest, SVM, K-NN, and XGBoost. The final step involves attributing synthetic images to their respective source datasets (CelebA or FFHQ).}
    \label{fig:pipeline}
\end{figure}

The study focuses on facial images manipulated or generated by GAN architectures trained on CelebA \cite{liu2015deep} or FFHQ\footnote{https://github.com/NVlabs/ffhq-dataset}. Synthetic data were produced using five widely used GAN models: AttGAN \cite{he2019attgan}, GDWCT \cite{cho2019image} and StarGAN \cite{choi2018stargan} (trained on CelebA), as well as StyleGAN \cite{karras2019style} and StyleGAN2 \cite{karras2020analyzing} (trained on FFHQ). Each dataset was divided into training (for a total of 56,000 images, 8,000 for each involved data type), validation (for a total of 8,000 images, 1,000 for each involved data type), and test (for a total of 8,000 images, 1,000 for each involved data type) sets. Our results show that the generated images have distinctive traces attributable to both the specific architecture and the used training data.

\begin{figure*}[t!]
  \centering
  \includegraphics[width=\linewidth]{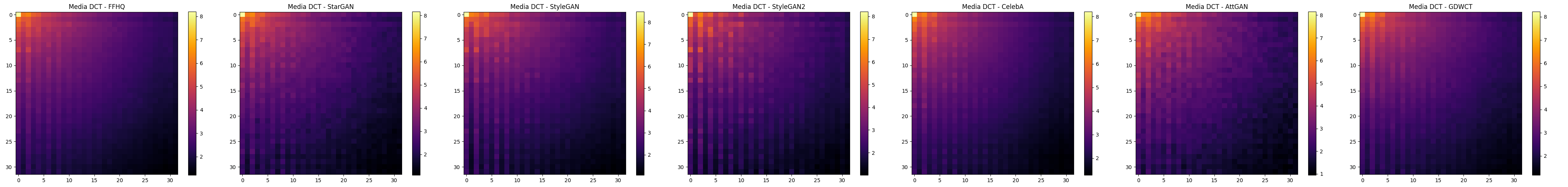}
  \caption{Average DCT coefficient heatmaps for real and GAN-generated images. Notable energy dispersion and spatial artifacts distinguish real (CelebA, FFHQ) from synthetic distributions (GDWCT, AttGAN, StarGAN, StyleGAN, StyleGAN2).}
  \label{fig:dct_heatmaps}
\end{figure*}
\begin{figure*}[t!]
  \centering
  \includegraphics[width=\linewidth]{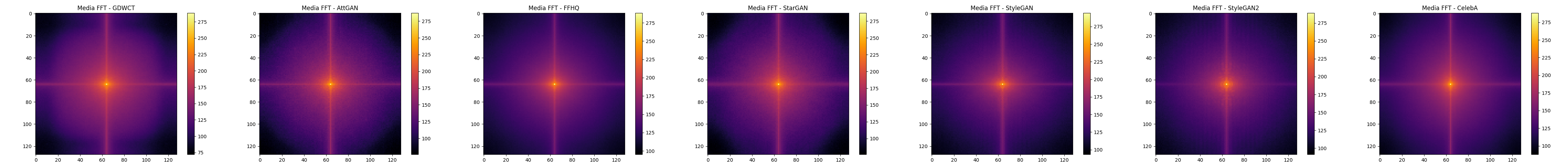}
  \caption{Mean FFT power spectra for real and GAN-generated images. Real datasets (CelebA, FFHQ) exhibit symmetrical low-frequency concentration, while synthetic distributions (GDWCT, AttGAN, StarGAN, StyleGAN, StyleGAN2) display cross-shaped high-frequency components.}
  \label{fig:fft_heatmaps}
\end{figure*}

Following dataset construction, we extracted a suite of multimodal features\footnote{Multimodal features are descriptors derived from different data representations, such as the DCT, FTT, and color histograms, capturing both frequency and color information for a richer image representation.} aimed at capturing both global and local characteristics of the images. In the frequency domain, we applied DCT and FFT to convert images to their spectral representations. These transforms allowed us to identify dataset-specific periodic artifacts and compression-related signatures. Color information was captured using normalized RGB histograms, while textural and structural properties were modeled through SIFT descriptors. The extracted features from each modality were concatenated to form a single feature vector for each image. To mitigate scale discrepancies, all features were individually standardized using zero-mean and unit-variance normalization, ensuring consistent scaling across different feature types, including DCT, FFT, and color histograms.

We employed four supervised machine learning algorithms—Random Forest, Support Vector Machines (SVM) with a linear kernel, K-Nearest Neighbors (K-NN = 3), and XGBoost—for classification tasks. The primary objectives were: (1) distinguishing between real and synthetic images, and (2) attributing synthetic images to their respective source datasets. The performance of each classifier was evaluated in terms of accuracy, precision, recall, and F1-score. Figure~\ref{fig:pipeline} illustrates the methodological pipeline, from dataset selection to feature extraction and classification, culminating in the attribution of synthetic images to their source datasets (CelebA or FFHQ). 



\section{Experimental Results}
\label{sec:res}
\subsection{Binary Classification: Real vs Deepfake}
Our experimental evaluation begins with an analysis of binary classification performance across diverse feature sets and models, followed by an exploration of dataset-specific attribution capabilities. This dual focus demonstrates a promising forensic capability to analyze synthetic images generated by GANs, shifting the focus from detection to accurately identifying their source training datasets for addressing copyright and legal concerns. By leveraging statistical analysis, our classifiers, which utilize multi-modal features, achieved accuracies of 98-99\% and strong F1-scores in distinguishing images generated from different datasets within the training domain. This confirms the presence of distinct statistical signatures associated with specific datasets, such as CelebA and FFHQ, used to train known GAN architectures.

Additionally, our models demonstrated consistently high performance in the binary classification task of distinguishing real images from synthetic deepfakes across a variety of feature sets, as summarized in Table~\ref{tab:real_fake_classification}. These features include DCT coefficients, color histograms (RGB), frequency domain characteristics (FFT), and combined feature sets (RGB + DCT + FFT). 

\begin{table}[t!]
\centering
\caption{Binary Classification of Real vs Fake Images for Different Feature Sets}
\label{tab:real_fake_classification}
\renewcommand{\arraystretch}{1.2} 
\begin{IEEEeqnarraybox}[
\IEEEeqnarraystrutmode
\IEEEeqnarraystrutsizeadd{1pt}{1pt}]{l l c c c c}
\hline
\textbf{Feature} & \textbf{Model} & \textbf{Accuracy } & \textbf{Precision } & \textbf{Recall } & \textbf{F1-score} \\
\hline
\multirow{4}{*}{DCT} 
& KNN            & 0.8625 & 0.9231 & 0.8182 & 0.8675 \\
& SVM            & \textbf{0.9500} & \textbf{1.0000} & \textbf{0.9091} & \textbf{0.9524} \\
& RF             & 0.9000 & 0.9286 & 0.8864 & 0.9070 \\
& XGBoost        & 0.9000 & 0.9091 & \textbf{0.9091} & 0.9091 \\
\hline
\multirow{4}{*}{RGB} 
& KNN            & 0.6625 & \textbf{0.7429} & 0.5909 & 0.6582 \\
& SVM            & 0.6250 & 0.7188 & 0.5227 & 0.6053 \\
& RF             & \textbf{0.6750} & 0.7368 & \textbf{0.6364} & \textbf{0.6829} \\
& XGBoost        & 0.6375 & 0.6829 & \textbf{0.6364} & 0.6588 \\
\hline
\multirow{4}{*}{FFT} 
& KNN            & 0.5500 & 0.6250 & 0.4545 & 0.5263 \\
& SVM            & \textbf{0.6750} & \textbf{0.6957} & \textbf{0.7273} & \textbf{0.7111} \\
& RF             & 0.5625 & 0.6286 & 0.5000 & 0.5570 \\
& XGBoost        & 0.6375 & 0.6829 & 0.6364 & 0.6588 \\
\hline
\multirow{4}{*}{RGB+DCT+FFT} 
& KNN            & 0.8625 & 0.9231 & 0.8182 & 0.8675 \\
& SVM            & \textbf{0.9625} & \textbf{1.0000} & \textbf{0.9318} & \textbf{0.9647} \\
& RF             & 0.9000 & 0.9286 & 0.8864 & 0.9070 \\
& XGBoost        & 0.8875 & 0.9070 & 0.8864 & 0.8966 \\
\hline
\end{IEEEeqnarraybox}
\end{table}

Notably, SVM achieved the highest F1-score (0.9647) with the combined feature set (RGB + DCT + FFT), reflecting the discriminative power of this multi-modal approach. 

\subsection{Dource Dataset Attribution}

As the features demonstrate a strong ability to discriminate synthetic content, this preliminary study explores their potential to analyze and infer the training dataset employed.

\begin{figure*}[t!]
  \centering
  \includegraphics[width=\linewidth]{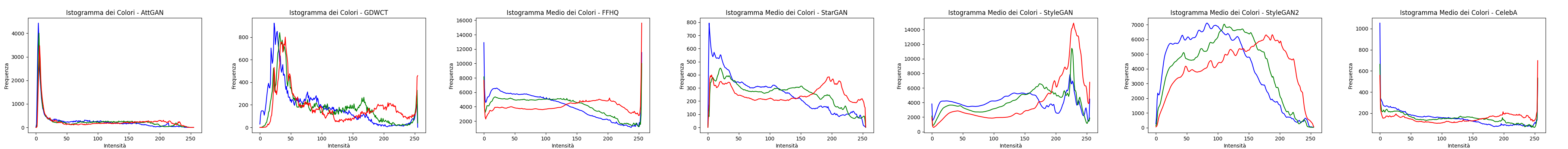}
  \caption{Average RGB color histograms. Real datasets maintain characteristic channel skews, while GAN-generated images display color balancing patterns.}
  \label{fig:rgb_histograms}
\end{figure*}
As illustrated in Figure~\ref{fig:dct_heatmaps}, DCT representations exhibit distinct characteristics across datasets. Real datasets, such as CelebA and FFHQ, demonstrate smoother frequency decay, reflecting natural image statistics. In contrast, GAN-generated images display localized high-frequency artifacts, indicative of synthesis-related transformations. These DCT-based features alone achieved performance comparable to the full feature set, highlighting their critical role in dataset attribution. Similarly, as shown in Figure~\ref{fig:fft_heatmaps}, FFT power spectra reveal clear distinctions between real and synthetic images. Real datasets (CelebA, FFHQ) exhibit centralized, isotropic low-frequency energy concentration, consistent with natural image properties. Conversely, GAN-generated images show cross-shaped spectral artifacts, likely arising from convolutional upsampling patterns and checkerboard effects, further aiding in dataset attribution.

To quantitatively assess the discriminative power of the extracted features, we trained the previously mentioned supervised classifiers. Performance was assessed on different feature sets individually and in combination, as summarized in Tables~\ref{tab:color_features}--\ref{tab:combined_features}.

\begin{itemize}
    \item Color-based features (HSV histograms, Table \ref{tab:color_features}) yielded solid results, with XGBoost achieving an accuracy of 90.2\% and F1-score of 0.9013. 
    \item Fourier-based features (Table~\ref{tab:fft_features}) provided even better performance, with both SVM and XGBoost surpassing 94\% accuracy. These features appear to encode critical information about image frequency content and structural regularity, which are often affected by GAN-specific generation artifacts.
    \item DCT features (Table~\ref{tab:dct_features}) performed slightly less robustly than Fourier ones but still demonstrated their utility, especially when used with ensemble classifiers such as XGBoost (83.3\% accuracy). The drop in recall and F1-score for simpler models indicates that DCT-based information may require more sophisticated learners to be fully exploited.
\end{itemize}

When combining DCT and FFT features, classification performance significantly improved across all models. XGBoost, in particular, achieved the highest overall performance, reaching 95.5\% accuracy and an F1-score of 0.9544. These results highlight the complementary nature of global frequency transforms in capturing dataset-specific statistical cues embedded in deepfake images.

\begin{table}[H]
\centering
\caption{Classification using color features (HSV)}
\label{tab:color_features}
\begin{tabular}{lcccc}
\hline
\textbf{Model} & \textbf{Accuracy} & \textbf{Precision} & \textbf{Recall} & \textbf{F1-score} \\
\hline
Random Forest & 0.870 & 0.8749 & 0.870 & 0.8677 \\
SVM           & 0.882 & 0.8827 & 0.882 & 0.8817 \\
KNN           & 0.757 & 0.7617 & 0.757 & 0.7405 \\
XGBoost       & \textbf{0.902} & \textbf{0.9039} & \textbf{0.902} & \textbf{0.9013} \\
\hline
\end{tabular}
\end{table}

\begin{table}[H]
\centering
\caption{Classification using Fourier-based features}
\label{tab:fft_features}
\begin{tabular}{lcccc}
\hline
\textbf{Model} & \textbf{Accuracy} & \textbf{Precision} & \textbf{Recall} & \textbf{F1-score} \\
\hline
Random Forest & 0.923 & 0.9299 & 0.923 & 0.9207 \\
SVM           & \textbf{0.943} & \textbf{0.9431} & \textbf{0.943} & \textbf{0.9426} \\
KNN           & 0.925 & 0.9352 & 0.925 & 0.9194 \\
XGBoost       & 0.940 & 0.9408 & 0.940 & 0.9394 \\
\hline
\end{tabular}
\end{table}

\begin{table}[H]
\centering
\caption{Classification using DCT-based features}
\label{tab:dct_features}
\begin{tabular}{lcccc}
\hline
\textbf{Model} & \textbf{Accuracy} & \textbf{Precision} & \textbf{Recall} & \textbf{F1-score} \\
\hline
Random Forest & 0.792 & \textbf{0.8859 }& 0.792 & 0.7296 \\
SVM           & 0.682 & 0.6918 & 0.682 & 0.6811 \\
KNN           & 0.682 & 0.7241 & 0.682 & 0.6857 \\
XGBoost       & \textbf{0.833} & 0.8346 & \textbf{0.833} & \textbf{0.8309} \\
\hline
\end{tabular}
\end{table}

\begin{table}[H]
\centering
\caption{Classification using combined features (DCT + FFT)}
\label{tab:combined_features}
\begin{tabular}{lcccc}
\hline
\textbf{Model} & \textbf{Accuracy} & \textbf{Precision} & \textbf{Recall} & \textbf{F1-score} \\
\hline
Random Forest & 0.935 & 0.9400 & 0.935 & 0.9313 \\
SVM           & 0.945 & 0.9443 & 0.945 & 0.9442 \\
KNN           & 0.735 & 0.7612 & 0.735 & 0.7388 \\
XGBoost       & \textbf{0.955} & \textbf{0.9558} & \textbf{0.955} & \textbf{0.9544} \\
\hline
\end{tabular}
\end{table}


As shown in Figure~\ref{fig:rgb_histograms}, real datasets exhibit distinct RGB intensity distributions, with characteristic peaks and troughs across channels. In contrast, GAN-generated images tend to smooth and normalize the color profiles, revealing implicit regularization strategies employed during training.



The results show that this preliminary approach contributes to advancements in media authentication and provenance tracking by providing interpretable statistical signals regarding an image’s dataset origin. These signals can support forensic investigations and inform the development of regulatory frameworks aimed at protecting data rights and promoting ethical AI practices.


\section{Legal and Ethical Considerations}
\label{sec:legal}
The ability to trace synthetic content back to its training dataset carries significant ethical and legal implications \cite{RomeroMoreno_2024}. Identifying whether a deepfake was created using a public dataset (like CelebA) or a more restricted one (like FFHQ) can influence assessments of its legality \cite{AI-generateddeepfakes}. In forensic and legal contexts, dataset attribution is critical—especially when models are trained on copyrighted or personal data obtained without consent \cite{Firm}.
If a deepfake is traced to real individuals’ faces collected without permission, it may breach data protection laws such as the EU’s General Data Protection Regulation (GDPR) \cite{RomeroMoreno_2024}. In the U.S., legal claims often invoke the right of publicity, protecting individuals from the unauthorized commercial use of their likeness. A striking case is non-consensual deepfake pornography, which disproportionately targets women, as seen in the Reddit deepfake pornography ban in 2018 \cite{Agarwal_2018}.
The recently adopted AI Act by the European Union addresses these concerns by requiring transparency for generative AI models. Providers must disclose when content has been AI-generated, helping to prevent misuse and improving dataset traceability \cite{RomeroMoreno_2024}. The Act also strengthens protections around biometric and personal data, banning their use in high-risk models without explicit consent.
From an intellectual property perspective, many deepfakes fall into a legal grey area. Under current U.S. copyright law, only human-authored works are eligible for protection, leaving AI-generated content potentially unprotected. However, legal scholars have proposed recognizing the role of the human developer or commissioner as the ``author" of AI-generated works \cite{RomeroMoreno_2024}. This could align with the AI Act’s broader aim of ensuring safe, transparent, and accountable AI usage.
Several landmark cases underscore the legal and ethical quandaries of AI‑generated media:
\begin{itemize}
\item \textbf{Reddit Deepfake Porn Ban (2018)} – In late 2017, a subreddit hosting non‑consensual, face‑swapped porn videos grew to nearly 100 000 members before Reddit shut it down in February 2018 for policy violations \cite{Agarwal_2018}. The uproar prompted platforms to ban such content and influenced California’s AB 602, which criminalizes explicit deepfakes created without consent.
\item \textbf{Artists’ Copyright Suit vs. AI (2023)} – In March 2023, Sarah Andersen, Kelly McKernan, and Karla Ortiz led a class action against Stability AI, DeviantArt, and Midjourney, alleging their models were trained on millions of unlicensed artworks, including the plaintiffs’ own \cite{Vincent_2023}. A federal judge found the copyright claim plausible, even absent identical outputs, spotlighting whether scraping copyrighted work for AI training itself infringes.
\item \textbf{“Heart on My Sleeve” Deepfake Song (2023)} – In April 2023, an AI‑generated track mimicking Drake and The Weeknd went live on streaming services. Universal Music Group swiftly issued takedown notices, arguing the song infringed both copyright and personality rights \cite{Drake}. This episode blurred the lines between intellectual property and digital identity.
\end{itemize}

\section{Discussion}
Our experimental results demonstrate that statistical features extracted from synthetic images—such as DCT coefficients, FFT spectra, and color histograms—carry dataset-specific fingerprints that enable robust source attribution. This aligns with findings by Verdoliva \cite{Verdoliva2020Media}, who emphasized frequency-domain artifacts for forensic analysis, and Corvi et al. \cite{corvi2024continuous}, who explored spectral fingerprints for diffusion models. However, our work uniquely combines global frequency and local textural features across diverse GAN architectures, achieving superior classification accuracy (95.5\% with XGBoost) while maintaining interpretability. 

Compared to prior methods, our approach offers distinct advantages. Traditional techniques like PRNU \cite{lukas2006} or image hashing \cite{venkatesan2000} focus on real-image forensics and fail for synthetic media, whereas our frequency-based features exploit GAN-specific generation artifacts (e.g., checkerboard patterns in DCT heatmaps; see Fig.~\ref{fig:dct_heatmaps}). Similarly, deep learning-based detectors \cite{afchar2018,yu2019} often lack explainability, while our handcrafted features provide actionable insights into dataset provenance. For instance, cosine similarity analysis confirms that GAN outputs cluster closely with their training datasets (e.g., CelebA-trained StarGAN vs. FFHQ-trained StyleGAN2), echoing Maini et al. \cite{maini2021dataset}, who proposed dataset inference via model membership queries but required access to the generator weights—a limitation our passive method avoids.

The legal and ethical implications of dataset tracing (detailed in Section~\ref{sec:legal}) further distinguish our work. By linking synthetic content to training data, our framework addresses critical gaps in synthetic media governance, such as identifying unauthorized use of copyrighted or personal datasets. This complements proactive fingerprinting approaches \cite{yu2021artificial} but applies broadly to legacy GANs without requiring model modification. However, challenges remain in cross-domain generalization, particularly for high-fidelity models like StyleGAN2, where synthetic images closely mimic real distributions. Future work could integrate transfer learning or adversarial training to improve robustness, as suggested by Guarnera et al. \cite{Guarnera2022DeepfakeStyleTransferMixture} for style-transfer pipelines.

\section{Conclusion and Future Work}
\label{sec:conclusion}
This paper presented a novel forensic framework for attributing the source dataset of deepfake images generated by GANs. Through a combination of spectral, color, and local descriptors, we demonstrated that GAN outputs retain identifiable traces of their training data. Experimental results showed high classification accuracy and confirmed the discriminative power of frequency-domain features in dataset attribution.

Despite the promising results, limitations in cross-domain generalization suggest the need for future work focusing on transfer learning, domain adaptation, and integration with neural fingerprinting methods. Additionally, we plan to extend this framework to diffusion-based generative models, which represent the next frontier in synthetic media.

Ultimately, our approach supports a broader forensic and legal toolkit for navigating the challenges of AI-generated content, offering a scalable, interpretable method for establishing the provenance of synthetic images.

\section*{Acknowledgment}

This study has been partially supported by SERICS (PE00000014) under the MUR National Recovery and Resilience Plan funded by the European Union - NextGenerationE.
\balance{
\bibliographystyle{IEEEtran}
\bibliography{bibliography}

\begin{thebibliography}{10}
\providecommand{\url}[1]{#1}
\csname url@samestyle\endcsname
\providecommand{\newblock}{\relax}
\providecommand{\bibinfo}[2]{#2}
\providecommand{\BIBentrySTDinterwordspacing}{\spaceskip=0pt\relax}
\providecommand{\BIBentryALTinterwordstretchfactor}{4}
\providecommand{\BIBentryALTinterwordspacing}{\spaceskip=\fontdimen2\font plus
\BIBentryALTinterwordstretchfactor\fontdimen3\font minus \fontdimen4\font\relax}
\providecommand{\BIBforeignlanguage}[2]{{%
\expandafter\ifx\csname l@#1\endcsname\relax
\typeout{** WARNING: IEEEtran.bst: No hyphenation pattern has been}%
\typeout{** loaded for the language `#1'. Using the pattern for}%
\typeout{** the default language instead.}%
\else
\language=\csname l@#1\endcsname
\fi
#2}}
\providecommand{\BIBdecl}{\relax}
\BIBdecl

\bibitem{Whittaker2020All}
L.~Whittaker, T.~C. Kietzmann, J.~H. Kietzmann, and A.~Dabirian, ``{"All Around Me Are Synthetic Faces": The Mad World of AI-Generated Media},'' \emph{IT Professional}, vol.~22, pp. 90--99, 2020.

\bibitem{Zia2024Improving}
R.~Zia, M.~Rehman, A.~Hussain, S.~Nazeer, and M.~Anjum, ``{Improving synthetic media generation and detection using generative adversarial networks},'' \emph{PeerJ Computer Science}, vol.~10, 2024.

\bibitem{goodfellow2014generative}
I.~J. Goodfellow, J.~Pouget-Abadie, M.~Mirza, B.~Xu, D.~Warde-Farley, S.~Ozair, A.~Courville, and Y.~Bengio, ``{Generative adversarial nets},'' \emph{Advances in Neural Information Processing Systems}, vol.~27, 2014.

\bibitem{brown2020}
T.~B. Brown, B.~Mann, N.~Ryder, M.~Subbiah, J.~Kaplan, P.~Dhariwal, A.~Neelakantan, P.~Shyam, G.~Sastry, A.~Askell, S.~Agarwal, A.~Herbert-Voss, G.~Krueger, T.~Henighan, R.~Child, A.~Ramesh, D.~M. Ziegler, J.~Wu, C.~Winter, C.~Hesse, M.~Chen, E.~Sigler, M.~Litwin, S.~Gray, B.~Chess, J.~Clark, C.~Berner, S.~McCandlish, A.~Radford, I.~Sutskever, and D.~Amodei, ``{Language Models are Few-Shot Learners},'' 2020.

\bibitem{crawford2021}
K.~Crawford and T.~Paglen, ``{Excavating AI: The Politics of Training Data},'' 2021.

\bibitem{amerini2025}
I.~Amerini, M.~Barni, S.~Battiato, P.~Bestagini, G.~Boato, V.~Bruni, R.~Caldelli, F.~De~Natale, R.~De~Nicola, L.~Guarnera, S.~Mandelli, T.~Majid, G.~L. Marcialis, M.~Micheletto, A.~Montibeller, G.~Orrù, A.~Ortis, P.~Perazzo, G.~Puglisi, N.~Purnekar, D.~Salvi, S.~Tubaro, M.~Villari, and D.~Vitulano, ``{Deepfake Media Forensics: Status and Future Challenges},'' \emph{Journal of Imaging}, vol.~11, no.~3, p.~73, 2025.

\bibitem{Casu2024GenAI}
M.~Casu, L.~Guarnera, P.~Caponnetto, and S.~Battiato, ``{GenAI Mirage: The Impostor Bias and the Deepfake Detection Challenge in the Era of Artificial Illusions},'' \emph{Forensic Science International: Digital Investigation}, vol.~50, p. 301795, 2024.

\bibitem{Mirsky2020The}
Y.~Mirsky and W.~Lee, ``{The Creation and Detection of Deepfakes},'' \emph{ACM Computing Surveys}, vol.~54, pp. 1--41, 2020.

\bibitem{Ghodke2024Ethical}
S.~Ghodke, ``{Ethical Implications of Deepfake Technology},'' \emph{International Journal For Multidisciplinary Research}, 2024.

\bibitem{Verdoliva2020Media}
L.~Verdoliva, ``{Media Forensics and DeepFakes: An Overview},'' \emph{IEEE Journal of Selected Topics in Signal Processing}, vol.~14, pp. 910--932, 2020.

\bibitem{Corvi2022On}
R.~Corvi, D.~Cozzolino, G.~Zingarini, G.~Poggi, K.~Nagano, and L.~Verdoliva, ``{On The Detection of Synthetic Images Generated by Diffusion Models},'' in \emph{ICASSP 2023 - 2023 IEEE International Conference on Acoustics, Speech and Signal Processing (ICASSP)}, 2022, pp. 1--5.

\bibitem{Pontorno2024DeepFeatureX}
O.~Pontorno, L.~Guarnera, and S.~Battiato, ``{DeepFeatureX Net: Deep Features eXtractors Based Network for Discriminating Synthetic from Real Images},'' in \emph{Lecture Notes in Computer Science}, ser. International Conference on Pattern Recognition (ICPR 2024).\hskip 1em plus 0.5em minus 0.4em\relax Springer, 2024, vol. 15321, pp. 177--193.

\bibitem{Guarnera2022DeepfakeStyleTransferMixture}
L.~Guarnera, O.~Giudice, and S.~Battiato, ``{Deepfake Style Transfer Mixture: A First Forensic Ballistics Study on Synthetic Images},'' in \emph{Image Analysis and Processing -- ICIAP 2022}, ser. Lecture Notes in Computer Science, S.~Sclaroff, C.~Distante, M.~Leo, G.~M. Farinella, and F.~Tombari, Eds.\hskip 1em plus 0.5em minus 0.4em\relax Springer, 2022, vol. 13232, pp. 151--163.

\bibitem{khoo2022deepfake}
E.~Y.~W. Khoo, H.~Fang, and M.~C. Stamm, ``{Deepfake attribution: On the source identification of artificially generated images},'' \emph{WIREs Data Mining and Knowledge Discovery}, vol.~12, no.~3, p. e1438, 2022.

\bibitem{li2018ictu}
Y.~Li, M.-C. Chang, and S.~Lyu, ``{In ictu oculi: Exposing AI generated fake face videos by detecting eye blinking},'' \emph{arXiv preprint arXiv:1806.02877}, 2018.

\bibitem{jia2024}
S.~Jia \emph{et~al.}, ``{Exposing Lip-syncing Deepfakes from Mouth Inconsistencies},'' 2024.

\bibitem{hernandez2020}
J.~Hernandez-Ortega, R.~Tolosana, J.~Fierrez, and A.~Morales, ``{DeepFakesON-Phys: DeepFakes Detection based on Heart Rate Estimation},'' 2020.

\bibitem{liu2021}
H.~Liu, X.~Li, W.~Zhou, Y.~Chen, Y.~He, H.~Xue, W.~Zhang, and N.~Yu, ``{Spatial-Phase Shallow Learning: Rethinking Face Forgery Detection in Frequency Domain},'' in \emph{2021 IEEE/CVF Conference on Computer Vision and Pattern Recognition (CVPR)}, 2021, pp. 760--769.

\bibitem{afchar2018}
D.~Afchar, V.~Nozick, J.~Yamagishi, and I.~Echizen, ``{MesoNet: a Compact Facial Video Forgery Detection Network},'' in \emph{2018 IEEE International Workshop on Information Forensics and Security (WIFS)}, 2018, pp. 1--7.

\bibitem{yu2019}
N.~Yu, L.~S. Davis, and M.~Fritz, ``{Attributing Fake Images to GANs: Learning and Analyzing GAN Fingerprints},'' in \emph{2019 IEEE/CVF International Conference on Computer Vision (ICCV)}, 2019, pp. 7555--7565.

\bibitem{corvi2024continuous}
R.~Corvi, D.~Cozzolino, G.~Poggi, K.~Nagano, and L.~Verdoliva, ``{Continuous fake media detection: adapting deepfake detectors to new generative techniques},'' \emph{arXiv preprint arXiv:2406.08171}, 2024.

\bibitem{yang2022deepfake}
T.~Yang, Z.~Huang, J.~Cao, L.~Li, and X.~Li, ``{Deepfake Network Architecture Attribution},'' in \emph{Proceedings of the AAAI Conference on Artificial Intelligence}, vol.~36, no.~4, 2022, pp. 4662--4670.

\bibitem{guarnera2024mastering}
L.~Guarnera, O.~Giudice, and S.~Battiato, ``{Mastering deepfake detection: A cutting-edge approach to distinguish GAN and diffusion-model images},'' \emph{ACM Transactions on Multimedia Computing, Communications and Applications}, vol.~20, no.~11, pp. 1--24, 2024.

\bibitem{yu2021artificial}
N.~Yu, V.~Skripniuk, D.~Chen, and L.~S. Davis, ``{Artificial fingerprinting for generative models: Rooting deepfake attribution in training data},'' in \emph{Proceedings of the IEEE/CVF International Conference on Computer Vision}, 2021, pp. 14\,448--14\,457.

\bibitem{maini2021dataset}
P.~Maini, M.~Yaghini, and N.~Papernot, ``{Dataset inference: Ownership resolution in machine learning},'' \emph{arXiv preprint arXiv:2104.10706}, 2021.

\bibitem{lukas2006}
J.~Lukas, J.~Fridrich, and M.~Goljan, ``{Digital Camera Identification From Sensor Pattern Noise},'' \emph{IEEE Transactions on Information Forensics and Security}, vol.~1, no.~2, pp. 205--214, 2006.

\bibitem{venkatesan2000}
R.~Venkatesan, S.-M. Koon, M.~H. Jakubowski, and P.~Moulin, ``{Robust Image Hashing},'' \emph{IEEE Signal Processing Letters}, vol.~7, no.~12, pp. 347--349, 2000.

\bibitem{cozzolino2021}
D.~Cozzolino and L.~Verdoliva, ``{Noiseprint: A CNN-Based Camera Model Fingerprint},'' \emph{IEEE Transactions on Information Forensics and Security}, vol.~16, pp. 144--159, 2021.

\bibitem{frank2020}
J.~Frank, M.~Eisenberger, and J.~L. Schönberger, ``{Frequency Domain Analysis for Detecting GAN-Generated Images},'' in \emph{2020 IEEE/CVF Conference on Computer Vision and Pattern Recognition Workshops (CVPRW)}, 2020, pp. 1996--2005.

\bibitem{wang2022}
T.~Wang, X.~Zhang, and J.~Wang, ``{Deepfake Detection Using Color Space Analysis},'' in \emph{2022 IEEE International Conference on Multimedia and Expo (ICME)}, 2022, pp. 1--6.

\bibitem{yang2019}
J.~Yang, C.-T. Li, and R.~Caldelli, ``{Local Feature Analysis for Deepfake Detection},'' in \emph{2019 IEEE International Conference on Image Processing (ICIP)}, 2019, pp. 4235--4239.

\bibitem{liu2015deep}
Z.~Liu, P.~Luo, X.~Wang, and X.~Tang, ``{Deep Learning Face Attributes in the Wild},'' in \emph{Proceedings of the IEEE International Conference on Computer Vision}, 2015, pp. 3730--3738.

\bibitem{he2019attgan}
Z.~He, W.~Zuo, M.~Kan, S.~Shan, and X.~Chen, ``{AttGAN: Facial Attribute Editing by Only Changing What You Want},'' \emph{IEEE Transactions on Image Processing}, vol.~28, no.~11, pp. 5464--5478, 2019.

\bibitem{cho2019image}
W.~Cho, S.~Choi, D.~K. Park, I.~Shin, and J.~Choo, ``{Image-to-Image Translation via Group-Wise Deep Whitening-and-Coloring Transformation},'' in \emph{Proceedings of the IEEE/CVF Conference on Computer Vision and Pattern Recognition}, 2019, pp. 10\,639--10\,647.

\bibitem{choi2018stargan}
Y.~Choi, M.~Choi, M.~Kim, J.-W. Ha, S.~Kim, and J.~Choo, ``{StarGAN: Unified Generative Adversarial Networks for Multi-Domain Image-to-image Translation},'' in \emph{Proceedings of the IEEE Conference on Computer Vision and Pattern Recognition}, 2018, pp. 8789--8797.

\bibitem{karras2019style}
T.~Karras, S.~Laine, and T.~Aila, ``{A Style-based Generator Architecture for Generative Adversarial Networks},'' in \emph{Proceedings of the IEEE/CVF Conference on Computer Vision and Pattern Recognition}, 2019, pp. 4401--4410.

\bibitem{karras2020analyzing}
T.~Karras, S.~Laine, M.~Aittala, J.~Hellsten, J.~Lehtinen, and T.~Aila, ``{Analyzing and Improving the Image Quality of StyleGAN},'' in \emph{Proceedings of the IEEE/CVF Conference on Computer Vision and Pattern Recognition}, 2020, pp. 8110--8119.

\bibitem{RomeroMoreno_2024}
F.~Romero~Moreno, ``{Generative AI and deepfakes: a human rights approach to tackling harmful content},'' \emph{International Review of Law, Computers \& Technology}, vol.~38, no.~3, pp. 297--326, Sep. 2024.

\bibitem{AI-generateddeepfakes}
\BIBentryALTinterwordspacing
Rouse, ``\BIBforeignlanguage{en}{{AI-generated deepfakes: what does the law say?}}'' Sep. 2024. [Online]. Available: \url{https://rouse.com/insights/news/2024/ai-generated-deepfakes-what-does-the-law-say/}
\BIBentrySTDinterwordspacing

\bibitem{Firm}
\BIBentryALTinterwordspacing
{National Security Law Firm}, ``{Understanding the Laws Surrounding AI-Generated Images: Protecting Yourself Against Deepfakes and Other Harmful AI Content}.'' [Online]. Available: \url{https://www.nationalsecuritylawfirm.com/understanding-the-laws-surrounding-ai-generated-images-protecting-yourself-against-deepfakes-and-other-harmful-ai-content/}
\BIBentrySTDinterwordspacing

\bibitem{Agarwal_2018}
\BIBentryALTinterwordspacing
R.~Agarwal, ``{Reddit Bans Deepfakes: Subreddits with AI-Generated Videos, Celebrity Porn Removed},'' Feb. 2018. [Online]. Available: \url{https://beebom.com/reddit-bans-deepfakes-celebrity-porn/}
\BIBentrySTDinterwordspacing

\bibitem{Vincent_2023}
\BIBentryALTinterwordspacing
J.~Vincent, ``{AI art tools Stable Diffusion and Midjourney targeted with copyright lawsuit},'' Jan. 2023. [Online]. Available: \url{https://www.theverge.com/2023/1/16/23557098/generative-ai-art-copyright-legal-lawsuit-stable-diffusion-midjourney-deviantart}
\BIBentrySTDinterwordspacing

\bibitem{Drake}
\BIBentryALTinterwordspacing
T.~FADER, ``\BIBforeignlanguage{en}{{Drake/The Weeknd deepfake song “Heart on My Sleeve” submitted to Grammys}},'' Sep. 2023. [Online]. Available: \url{https://www.thefader.com/2023/09/06/drake-the-weeknd-song-heart-on-my-sleeve-submitted-to-grammys}
\BIBentrySTDinterwordspacing

\end{thebibliography}
}
\end{document}